\documentclass[letterpaper, 10 pt, conference]{ieeeconf}  

\IEEEoverridecommandlockouts                              
\usepackage{amsfonts}
\usepackage{mathtools}
\usepackage{amsmath}
\usepackage{amssymb}
\usepackage{framed}
\usepackage{balance}
\usepackage{subfigure}
\usepackage{graphics} 
\usepackage{lipsum}
\usepackage{graphicx}
\usepackage{mathptmx} 
\usepackage{fancyhdr}
\usepackage[ruled,vlined]{algorithm2e}
\usepackage{setspace}
\usepackage{amssymb}  
\usepackage{algpseudocode}
\usepackage{booktabs}
\usepackage{tabularx}
\usepackage{booktabs}
\usepackage{array}
\usepackage{etex}
\usepackage{adjustbox}
\usepackage{adjustbox,lipsum}
\usepackage[table]{xcolor}
\usepackage{gensymb}
\usepackage{nicefrac}
\usepackage{mdwtab}
\usepackage{multirow}
\usepackage{colortbl}
\usepackage{hyperref}
\usepackage{color,soul}
\usepackage{epsfig} 
\usepackage{graphics} 

\DeclareMathAlphabet{\mathcal}{OMS}{cmsy}{m}{n}

\usepackage{stmaryrd}

\algrenewcommand\algorithmicforall{\textbf{foreach}}
\algrenewcommand\algorithmicindent{.8em}

\usepackage{accents}
\newcommand*{\dt}[1]{\accentset{\mbox{\bfseries .}}{#1}}
\newcommand*{\ddt}[1]{\accentset{\mbox{\bfseries .\hspace{-1.0ex}.}}{#1}}

\begin{document}

\title{\LARGE \bf
Human to Robot Whole-Body Motion Transfer
}
\author{Miguel Arduengo$^{1}$, Ana Arduengo$^{1}$, Adri\`a Colom\'e$^{1}$, Joan Lobo-Prat$^{1}$ and Carme Torras$^{1}$
\thanks{This work has been partially funded by the European Union Horizon 2020 Programme under grant agreement no. 741930 (CLOTHILDE) and by the Spanish State Research Agency through the María de Maeztu Seal of Excellence to IRI [MDM-2016-0656].}
\thanks{$^{1}$ Institut de Robòtica i Informàtica Industrial, CSIC-UPC (IRI), Barcelona. \{marduengo, aarduengo, acolome, jlobo, torras\}@iri.upc.edu}
}

\maketitle
\thispagestyle{empty}
\pagestyle{empty}



\begin{abstract}

Transferring human motion to a mobile robotic manipulator and ensuring safe physical human-robot interaction are crucial steps towards automating complex manipulation tasks in human-shared environments. In this work, we present a novel human to robot whole-body motion transfer framework. We propose a general solution to the correspondence problem, namely a mapping between the observed human posture and the robot one. For achieving real-time imitation and effective redundancy resolution, we use the whole-body control paradigm, proposing a specific task hierarchy, and present a differential drive control algorithm for the wheeled robot base. To ensure safe physical human-robot interaction, we propose a novel variable admittance controller that stably adapts the dynamics of the end-effector to switch between stiff and compliant behaviors. We validate our approach through several real-world experiments with the TIAGo robot. Results show effective real-time imitation and dynamic behavior adaptation. This constitutes an easy way for a non-expert to transfer a manipulation skill to an assistive robot.

\end{abstract} 

\section{INTRODUCTION}

\begin{figure}[t]
    \centering
    {\includegraphics[width=0.92\linewidth]{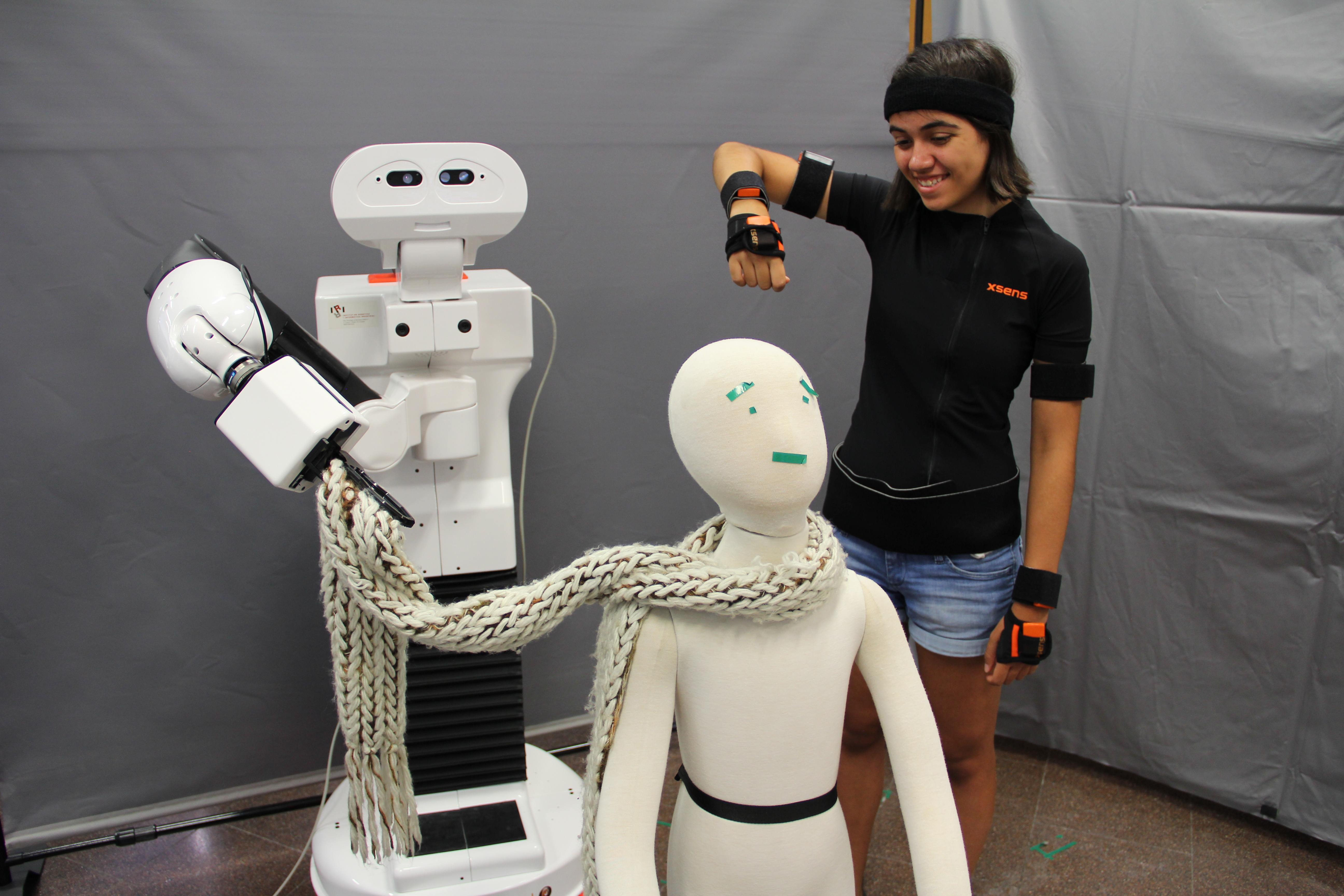}}
    \caption{Transferring human motion to robots while ensuring safe human-robot physical interaction can be a powerful and intuitive tool for teaching assistive tasks such as helping people with reduced mobility to get dressed.}
    \vspace{-1.8em}
    \label{Fig1}
\end{figure}

Service robots may assist people at home in the future. However, robotic systems still face several challenges in unstructured human-shared environments. One of the main challenges is to achieve human-like manipulation skills \cite{Torras2016}. Learning from demonstrations is arising as a promising paradigm in this regard \cite{Argall2009}\cite{Chernova2014}\cite{Hussein2017}. Rather than analytically decomposing and manually programming a desired behavior, a controller can be derived from observations of human performance. However, transferring human motion, in a way that demonstrations can be easily reproduced by the robot, ensuring a compliant and safe behavior when physically interacting with a person in assistive or cooperative domains, is not straightforward. Imitation is an intuitive whole-body motion transfer approach due to the similarities in embodiment between humans and service robots (Figure \ref{Fig1}). A fundamental problem is to create an appropriate mapping between the actions afforded to achieve corresponding states by the model and imitator agents \cite{Alissandrakis2006}. This problem is known in literature as the correspondence problem. It implies determining what and how to imitate. Another key challenge on the whole-body motion transfer problem, in human-centered robotics applications, is to ensure that the learned skill can be reproduced in a safe and compliant manner. This requires to adequately balance two opposite control objectives: a tight tracking of the motion being imitated, and a reactive behavior to interaction forces, allowing a steady-state position error. 

The classical approach for human motion transfer is kinesthetic teaching. The teacher holds the robot along the trajectories to be followed to accomplish a specific task, while the robot does gravity compensation \cite{Rozo2016}. The main advantage of this approach is that it avoids solving the correspondence problem. However, since the robot must be held, is not suitable for robots with a high number of degrees of freedom (DOF) such as humanoids, limiting the taught motions. By solving the human-robot correspondence, the robot can learn more human-like motion policies. This field has been very active in the recent years. In \cite{Koenemann2014} the authors present an approach for imitating the human hand and feet position with a Nao robot by solving the inverse kinematics (IK) of the robot, while ensuring stability with a balancing controller. Instead of solving the IK for the robot end-effectors, in \cite{Penco2018} the authors propose a direct joint angle retargeting, alleviating the computational complexity. The main limitation of these works is that their solutions are strongly robot-dependent. This issue is addressed in \cite{Darvish2019}, where the authors enhance scalability for motion transfer by establishing a correspondence between human and robot upper-body links in task space. Although excellent results are achieved regarding the motion imitation, physical interaction, which is of critical importance for robots operating in human environments, is not considered in these works. 

Variable admittance control \cite{Villani2008} is a suitable approach for modulating the robot behavior during physical human-robot interaction. However, research efforts have been mainly focused on cases where the robot motion is only driven by the force exerted by a human \cite{Keemink2017}\cite{Ferraguti2019}. This is not the case for a robot imitating the human posture, where a reference position should also be considered. An approach with a constant impedance profile has been proposed recently in \cite{Wu2019}.

In this work we propose a general framework for easily transferring human-like motion to a mobile robot manipulator. We propose a novel imitation based whole-body motion transfer interface. The main contributions are on the formulation of a general solution to the correspondence problem, and the definition of a control scheme for effective real-time whole-body imitation while ensuring compliance and stability during physical human-robot interaction. The paper is organized as follows: in Section \ref{motion_transfer} we discuss the main aspects of the proposed system; in Section \ref{experiments} we present the conducted experiments to validate our approach; finally in Section \ref{conclusion}, we summarize the main conclusions.

\vspace{-0.5mm}

\section{Whole-Body Motion Transfer Framework}
\label{motion_transfer}

\vspace{-0.5mm}

Achieving accurate, real-time robotic imitation of human motion while ensuring safe physical human-robot interaction, involves several steps. The first one is to adequately capture human motion \cite{Beth2003}\cite{Endres2012}. Then, we should determine what and how to imitate. This implies not only to determine a correspondence between human posture and the robot configuration, but also an effective management of the DOF and account for the robot constraints \cite{Stanton2012}\cite{Tunstel2013}. Finally, compliance and an accurate position control should be adequately balanced since they involve different dynamics \cite{Kronander2015}. In this section we present the proposed methods to address these challenges.

\vspace{-0.5mm}

\subsection{Motion Capture} 
\label{motion_capture}

\vspace{-0.5mm}

Motion capture is a way to digitally record human movements. Data is mapped on a digital model in 3D \cite{Nakazawa2018}. Inertial motion capture, compared to camera-based systems, does not rely on any external infrastructure allowing it to be used anywhere \cite{Schepers2018}. We use the Xsens MVN motion capture suit. It uses 17 body-attached inertial measurement units (IMUs) to obtain a body configuration and provide a real-time estimation of the human posture. The suit is supplied with MVN Studio software that processes raw sensor data and estimates body segment position and orientation. It is capable of sending real-time motion capture data of 23 body segments using the UDP/IP communication protocol.


\subsection{Correspondence Problem}
\label{correspondence}

\vspace{-0.5mm}

The correspondence problem can be stated as: given an observed behaviour of the human, which from a given starting posture evolves through a sequence of subgoals in poses, the robot must find and execute a sequence of actions using its own (possibly dissimilar) embodiment, which from a corresponding starting posture, leads through corresponding subgoals to corresponding poses \cite{Alissandrakis2006}. This accounts that the human and robot may not share the same morphology or affordances. This problem requires adequate considerations regarding the differences in the kinematic chains and joint limits. It can be divided into three subproblems:
\begin{itemize}
    \item \textit{Observation}: Measure the person state $\left(f_p^o:\mathcal{P}\rightarrow\mathcal{O}\right).$
     \item \textit{Equivalence}: Establish a relation between the observed state and the robot desired posture  $\left(f_o^g:\mathcal{O}\rightarrow\mathcal{G}\right)$.
     \item  \textit{Imitation}: Determine the robot configuration that allows to achieve the goal state $\left(f_r^g:\mathcal{R}\rightarrow\mathcal{G}\right)$.
     \vspace{-4mm}
\end{itemize} where $f^a_b$ is the mapping from $\mathcal{B}$ to $\mathcal{A}$ and $\mathcal{P}$, $\mathcal{O}$, $\mathcal{G}$ and $\mathcal{R}$ refer to the person state, observation, goal and robot configuration spaces respectively. 

Formally, the problem is finding $f_p^r: \mathcal{P}\rightarrow\mathcal{R}$, defined as: \begin{equation}
    f_p^r= f_p^o \circ f_o^g \circ \left(f_r^g\right)^{-1}
\end{equation} where $\circ$ is the composition operator and $()^{-1}$ the inverse. Based on this definition, the solution depends on the motion capture system, as it conditions the observation space. Using a system like the one presented in Section \ref{motion_capture}, $\small{\mathcal{O}=SE\left(3\right)^n}$, where $n$ is the number of observed person segments and $\small{SE\left(3\right)^n}$ an $n$-dimensional array of three-dimensional Euclidean groups. Each element of the array is a homogeneous transformation from reference $i$ to $j$: \begin{equation} \small \mathbf{T}_i^j=\left(\begin{array}{cc}
      \mathbf{R}_i^j &  \mathbf{p}_i^j\\
      \mathbf{0} & 1 
\end{array}\right)\end{equation} where $\small{\mathbf{R}_i^j \in SO(3)}$ and $\small{\mathbf{p}_i^j \in \mathbb{R}^3}$ are the rotational and translational components, respectively. 

Then, $\small{f_p^o:\mathcal{P}\rightarrow SE\left(3\right)^n}$ and $\small{f_o^g \circ \left(f_r^g\right)^{-1}: SE\left(3\right)^n\rightarrow\mathcal{R}}$. Ultimately, the problem is finding a mapping between Cartesian and robot configuration spaces. This is a problem widely studied in robotics. Currently, specially in humanoid robotics research, where robots have a high number of DOF, frameworks for defining $\small{\left(f_r^g\right)^{-1}:SE\left(3\right)^m\rightarrow\mathcal{R}}$ in such a way that the pose of $m$ robot links can be constrained in Cartesian space are being developed, such as Whole-Body Control \cite{Sentis2018}. Therefore, in order to make our proposed solution as general as possible, it seems convenient to define a correspondence function such as $\small{f_o^g:SE\left(3\right)^n\rightarrow SE\left(3\right)^m}$. This can be summarized in the following scheme:\begin{equation*}\small\underset{person}{\mathcal{P}}\xrightarrow{\;\;\;f_p^o\;\;\;}\underset{observations}{\mathcal{O}=SE\left(3\right)^n}\xrightarrow{\;\;\;f_o^g\;\;\;}\underset{goal}{\mathcal{G}=SE\left(3\right)^m}\xrightarrow{\;\;\left(f_r^g\right)^{-1}}\underset{robot}{\mathcal{R}}\end{equation*}

\begin{figure}[h]

\vspace{-2mm}

    \centering
    {\includegraphics[width=0.94\linewidth]{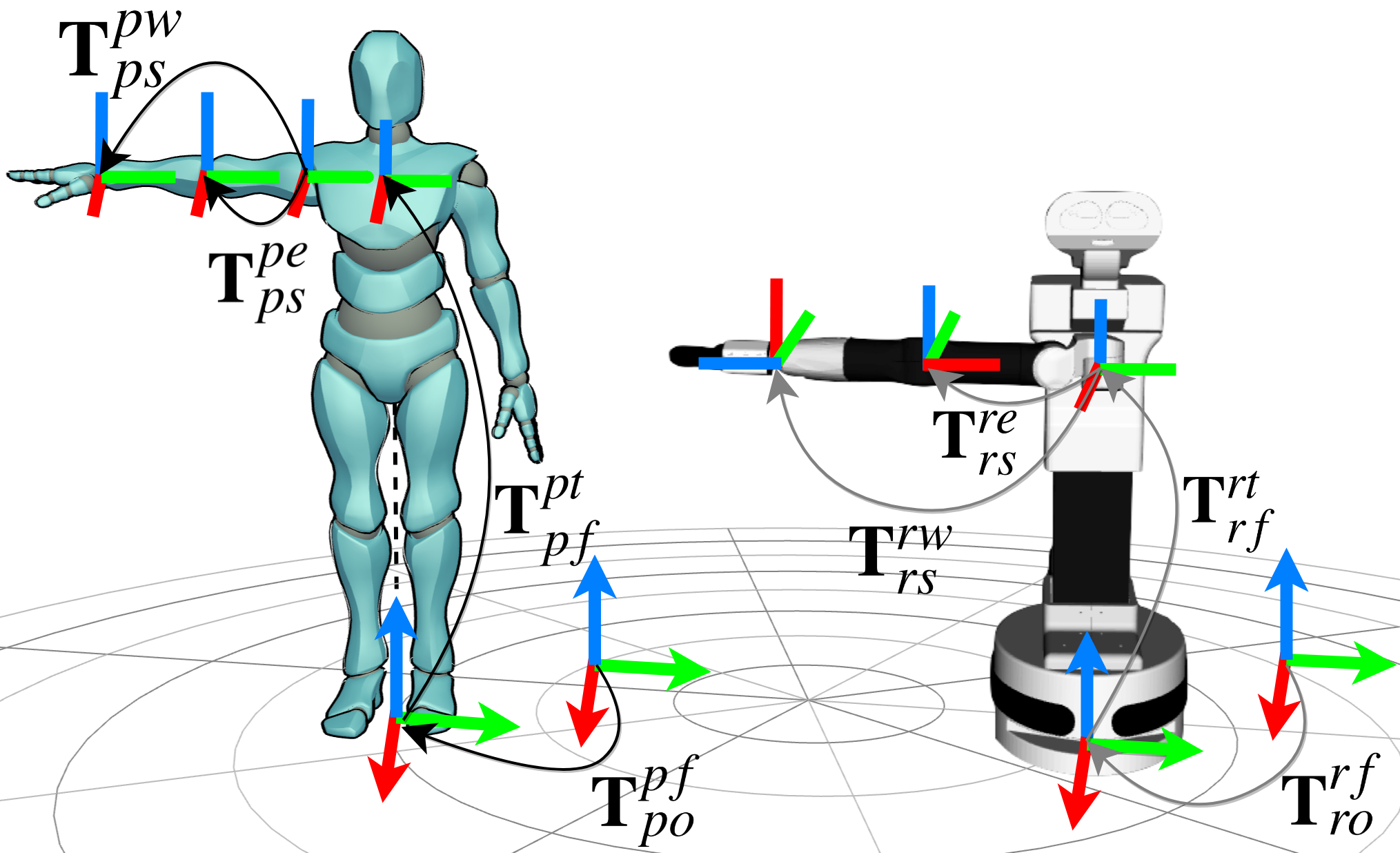}}
    \caption{Mapping in Cartesian space for an equivalent pose between a human model and the TIAGo robot. The colors red, green and blue are the x, y and z axes of each reference frame, respectively. Note that for the particular case of a robot with a morphology like TIAGo $rt\equiv rs$.}
    \label{Fig3}
    \vspace{-1.3em}
\end{figure}

We consider that the pose is equivalent if, with respect to an arbitrary fixed reference frame, the difference in position and orientation of the person's right (or left) wrist, elbow, chest, and the projection of the pelvis onto the floor; and the equivalent links of the robot up to an scaling factor, is zero. Given $\small{\mathbf{T}_{po}^{pf}}$, $\small{\mathbf{T}_{pf}^{pt}}$, $\small{\mathbf{T}_{ps}^{pe}}$ and $\small{\mathbf{T}_{ps}^{pw}}$ where $po$, $pf$, $pt$, $ps$, $pe$ and $pw$ stand for the person arbitrary origin, virtual footprint, torso, shoulder, elbow and wrist reference frames respectively and an example of an equivalent person-robot pose i.e. corresponding state (e.g. Figure \ref{Fig3}). We propose that the analogous robot pose is fully-determined by $\small{\mathbf{T}_{ro}^{rf}}$, $\small{\mathbf{T}_{rf}^{rt}}$, $\small{\mathbf{T}_{rs}^{re}}$ and $\small{\mathbf{T}_{rs}^{rw}}$, where $ro$, $rf$, $rt$, $rs$, $re$, $rw$ are the equivalent robot links. Their rotational components are defined as: \begin{equation}\small
\mathbf{R}_{ri}^{rj}=\mathbf{Rs}_{pi}^{ri} \cdot \mathbf{R}_{pi}^{pj}\end{equation} where $pi$ and $ri$ (also $pj$ and $rj$) are arbitrary equivalent person and robot links, and $\mathbf{Rs}$ is the rotation when both are in the equivalent example pose. The translational components are similarly defined as:

\vspace{-3mm}

\begin{equation}
\small\mathbf{p}_{ro}^{rf} = \mathbf{p}_{po}^{pf} \qquad \mathbf{p}_{rf}^{rt} = L_{rf}^{rt}\cdot\frac{\mathbf{p}_{pf}^{pt}}{\left\|\mathbf{p}_{pf}^{pt}\right\|}\end{equation}
\begin{equation}\newline\small\mathbf{p}_{rs}^{re}=L_{rs}^{re}\cdot\frac{\mathbf{p}_{ps}^{pe}}{\left\|\mathbf{p}_{ps}^{pe}\right\|} \qquad  \mathbf{p}_{rs}^{rw}=\mathbf{p}_{rs}^{re}+L_{re}^{rw}\cdot\frac{\mathbf{p}_{ps}^{pw}-\mathbf{p}_{ps}^{pe}}{\left\|\mathbf{p}_{ps}^{pw}-\mathbf{p}_{ps}^{pe}\right\|}\end{equation} where $L_{rf}^{rt}$ is the robot's base to torso height when the torso is fully extended, $L_{rs}^{re}$ and $L_{re}^{rw}$ are the lengths of the robot's equivalent shoulder to elbow and elbow to wrist segment respectively. Therefore, a complete definiton of $\small{f_o^g:SE\left(3\right)^n\rightarrow SE\left(3\right)^m}$ is provided.

\subsection{Whole Body Control}

Whole-Body Control (WBC) \cite{RAS-WBC2019} has been proposed as a promising research direction when using robots with many DOF and several simultaneous objectives. The redundant DOF can be conveniently exploited to meet the multiple tasks constraints \cite{Iskandar2019}\cite{Oh2019}. Given a set of $k$ control actions
targeting an individual task $\mathbf{x}_i\in SE(3)$, which defines a desired motion in Cartesian space, a generic definition of a WBC is \cite{Sentis2018}:\begin{equation}
\dt{\mathbf{q}}=\mathbf{J}_1^{\dagger}\dt{\mathbf{x}}_1+\mathbf{J}_2^{\dagger}\dt{\mathbf{x}}_2+\dots+\mathbf{J}_k^{\dagger}\dt{\mathbf{x}}_k\end{equation} where $\mathbf{q}\in\mathcal{R}$ is the robot configuration, $\mathbf{J}_i^{\dagger}=\mathbf{J}_i^T\left(\mathbf{J}_i\mathbf{J}_i^T\right)^{-1}$ is the Moore-Penrose pseudo-inverse of the $i^{th}$ task Jacobian which is defined by $\dt{\mathbf{x}}_i=\mathbf{J}_i\dt{\mathbf{q}}$. A task can represent, for example, the end-effector pose or the available joint range. 

A hierarchical ordering among tasks can be defined. Let $\mathbf{N_{A}}_i=\mathbf{I}-\mathbf{J_{A}}_i^{\dagger}\mathbf{J_{A}}_i$ be the null-space projector of the augmented Jacobian $\mathbf{J_A}_i=\left(\mathbf{J}_1, \dots, \mathbf{J}_i\right)$. Then, the joint velocity can be determined with the following relationship \cite{Siciliano1991}: \begin{equation}
\dt{\mathbf{q}}_i=\dt{\mathbf{q}}_{i-1}+\mathbf{J}_i\mathbf{N_A}_{i-1}\left(\dt{\mathbf{x}}_i-\mathbf{J}_i\dt{\mathbf{q}}_{i-1}\right)\qquad \dt{\mathbf{q}}_1=\mathbf{J}_1^{\dagger}\dt{\mathbf{x}}_1 \end{equation} This allows the $i^{th}$ task to be executed with lower priority with respect to the previous $i-1$ task, not interfering with the higher priority tasks. If $\mathbf{J}_i$ is singular, the $i^{th}$ task cannot be satisfied. However, the subsequent tasks are not affected since the dimension of the null-space of $\mathbf{J_A}_i$ is not decreased.

For whole-body imitation, the robot needs to achieve multiple varying goals in Cartesian space simultaneously. This makes WBC a suitable control framework, since they can be defined as a set of tasks with an adequate hierarchy. The main advantage over other inverse kinematic solvers is that the WBC can find online solutions automatically preventing self-collisions and ensuring joint limits. We are using PAL robotics implementation for the upper-body, which is based on the Stack of Tasks \cite{Mansard2009}. Taking into consideration the equivalence human-robot relations presented in the previous section, we propose the following task hierarchy:

\begin{enumerate}
    \item Joint limit avoidance
    \item Self-collision avoidance
    \item Torso position control
    \item End-effector pose admittance control
    \item Elbow pose control
\end{enumerate}

The first two tasks should always be active with the higher priority for safety reasons. The torso task is of higher priority because, by constraining the torso, the arm DOF are not affected, but the opposite is not true. Then, defining the end-effector task with the higher priority we ensure a correct end-effector goal tracking, which is important for manipulation tasks. The use of admittance control for this particular task in discussed in Section \ref{admittance}. Then, with the elbow task we ensure the arm posture imitation. In the particular case that the robot arm has human-like structure, which is the case of the TIAGo robot, a correct imitation can be achieved with the presented hierarchy. With the WBC we focus on redundancy resolution, finding the optimal configuration to accomplish the high-level task. 

\vspace{-3mm}

\subsection{Differential Drive Base Control}

Differential drive base is a mechanism used in many mobile robots, such as TIAGo or Roomba \cite{Morin2008}. It usually consists on two drive wheels mounted on a common axis  \cite{Topiwala2016}. Linear and angular velocity are the control commands \cite{Norr2017}. Let $\left(x, y, \theta\right)^T$ be the coordinates that define the base pose. Let $v$ and $\omega$ be the instantaneous linear and angular velocity commands respectively. The kinematic model is:\begin{equation}\small\left(\begin{array}{ccc}
     \dt{x} & \dt{y} & \dt{\theta}
\end{array}\right)=\left(\begin{array}{ccc}
     v\cos{\theta} & v\sin{\theta} & \omega
\end{array}\right)\end{equation} with the non-holonomic constraint $\dt{y}\cos{\theta}-\dt{x}\sin{\theta}=0$, which does not allow movements in the wheels' axis direction. 

Using the notation presented in Section \ref{correspondence}, the robot footprint pose should coincide with the person's, plus an arbitrary constant offset for a successful imitation. It is an inverse kinematics problem i.e., find the velocity commands that allow the robot to reach a given pose. Common path planning frameworks address this problem \cite{Gonzalez2016}. However, most of them are not suitable for cases where the goal is constantly changing at a rate of a human walking, which makes the robot remain in a planning state. Additionally, they do not consider backwards motion, which might be needed. We propose a computationally simple implementation, summarized in Algorithm \ref{algorithm1} to address these issues. When initialized, it assumes the person and the robot footprint frames are coincident in an arbitrary fixed reference frame. Then the relative transform between the person and the robot footprint is determined at each time step. When the robot is further than a certain margin to the reference, angular velocity commands orientate the robot towards the goal position. If the robot position is close enough, angular velocity commands align the robot with the goal orientation. 

\begin{algorithm}
\label{algorithm1}
\caption{Differential Drive Base Imitation}
\footnotesize 
\tcc{$\epsilon$,$\delta$,$\lambda$,$\sigma$: design parameters}
\tcc{$()_{yaw}$: yaw component of rotation}
\tcc{$()_{x,y}$: $x,y$ component of translation}
\tcc{$\mathbf{T}_a^b$: transforms defined in Section \ref{correspondence}}
\tcc{$x$ axis is assumed as forward}
\SetAlgoLined
$\mathbf{T}_{ro}^{po}=\mathbf{T}_{ro}^{rf}\mathbf{T}_{pf}^{po}$\tcp*{Initialize $\mathbf{T}_{rf}^{pf}=\mathbf{I}$}

\While {True} {
    $\mathbf{T}_{rf}^{pf}=\left(\mathbf{T}_{ro}^{pf}\right)^{-1}\mathbf{T}_{ro}^{po}\mathbf{T}_{po}^{pf}$\tcp*{Relative transform}
    
    \uIf {$\left\|\mathbf{p}_{rf}^{pf}\right\|<\epsilon$}{
        $v = 0 \qquad \omega = \lambda\, \cdot \left(\mathbf{R}_{rf}^{pf}\right)_{yaw}$
        }
    \uElseIf{$\left(\mathbf{p}_{rf}^{pf}\right)_{x}<0$ \textbf{and} $\Big|\left(\mathbf{R}_{rf}^{pf}\right)_{yaw}\Big|<\delta$} {
        $v = -\sigma\cdot\left\|\mathbf{p}_{rf}^{pf}\right\|$
            
        $\omega=\lambda \cdot\left(\arctan\frac{\left(\mathbf{p}_{rf}^{pf}\right)_{y}}{\left(\mathbf{p}_{rf}^{pf}\right)_{x}}-\pi\cdot sgn\left[\arctan\frac{\left(\mathbf{p}_{rf}^{pf}\right)_{y}}{\left(\mathbf{p}_{rf}^{pf}\right)_{x}}\right]\right)$
        }
    \Else{
        $v = \sigma \cdot \left\|\mathbf{p}_{rf}^{pf}\right\|\qquad\omega = \lambda\cdot\arctan\frac{\left(\mathbf{p}_{rf}^{pf}\right)_{y}}{\left(\mathbf{p}_{rf}^{pf}\right)_{x}}$
        }
 }
\end{algorithm}

\vspace{-5mm}
\subsection{Variable Admittance Control}
\label{admittance}

Admittance control \cite{Hogan1984} is a method where, by measuring the interaction forces, the set-point to a low-level motion controller is changed through a virtual spring-mass-damper model dynamics to achieve some preferred interaction responsive behavior \cite{Keemink2018}. In simple cases, the parameters of such a system can be identified in advance and kept fixed. However, when interaction forces are subject to uncertainties, the desired response can be adaptively regulated \cite{Peternel2016}. Variable admittance control allows to change the dynamics in a continuous manner during the task. When imitating the human posture in real-time, an accurate pose control is desirable, so a stiff behavior is preferable. On the other hand, when physically interacting with a human, a compliant (i.e. low stiffness) behavior is of vital importance to ensure safety \cite{Ott2015}\cite{Talignani2017}. The virtual end-effector dynamics: \begin{equation}\mathbf{M}\left(t\right)\ddt{\mathbf{e}}\left(t\right)+\mathbf{D}\left(t\right)\dt{\mathbf{e}}\left(t\right)+\mathbf{K}\left(t\right)\mathbf{e}\left(t\right)=\mathbf{F_{ext}}\left(t\right)
\end{equation} where inertia $\mathbf{M}\left(t\right)\in\mathbb{R}^{6\times 6}$, damping $\mathbf{D}\left(t\right)\in\mathbb{R}^{6\times 6}$ and stiffness $\mathbf{K}\left(t\right)\in\mathbb{R}^{6\times 6}$ determine the virtual dynamics of the robot, $\mathbf{e}\left(t\right)=\mathbf{x}\left(t\right)-\mathbf{x}_{ref}\left(t\right)\in\mathbb{R}^{6\times 1}$, when subjected to an external force $\mathbf{F_{ext}}\left(t\right)\in\mathbb{R}^{6\times 1}$. $\mathbf{x}_{ref}\left(t\right)$ and $\mathbf{x}\left(t\right)$ are, in our particular case, the reference position provided by the human and the position passed to the WBC, respectively.

If $\mathbf{M}\left(t\right)$, $\mathbf{D}\left(t\right)$ and $\mathbf{K}\left(t\right)$ are constant, the system will be asymptotically stable for any symmetric positive definite choice of the matrices. However, in this work we assume that $\mathbf{M}$ remains constant while $\mathbf{D}\left(t\right)$ and $\mathbf{K}\left(t\right)$ vary in time. It can be proved (see \cite{Kronander2016}) that for a constant, symmetric, positive definite $\mathbf{M}$, and $\mathbf{D}\left(t\right)$, $\mathbf{K}\left(t\right)$ continuously differentiable, the system is globally asymptotically stable if there exists a $\gamma>0$ such that, $\forall t \geq 0$:

\begin{enumerate}
    \item $\gamma\,\mathbf{M}-\mathbf{D}\left(t\right)$ is negative semidefinite
    \item $\dt{\mathbf{K}}\left(t\right)+\gamma\,\dt{\mathbf{D}}\left(t\right)-2\gamma\,\mathbf{K}\left(t\right)$ is negative definite
\end{enumerate}

Without loss of generality, we can assume that $\mathbf{M}$, $\mathbf{D}\left(t\right)$ and $\mathbf{K}\left(t\right)$ are diagonal matrices, since they can always be expressed in a suitable reference frame. Therefore, the system can be uncoupled in six independent scalar systems. To condense, we focus on the translational DOF. However, for the rotational components, the deduction is analogous:\begin{equation}m\,\ddt{e}\left(t\right)+d\left(t\right)\,\dt{e}\left(t\right)+k\left(t\right)\,e\left(t\right)=f_{ext}\left(t\right)\end{equation}
As design criteria, we will ensure a constant damping ratio $\zeta>0$. Thus, the damping is chosen as $d\left(t\right)=2\zeta\sqrt{m\,k\left(t\right)}$. By substituting on the second stability condition, it yields the following upper bound for the stiffness derivative:\begin{equation}\small\dt{k}\left(t\right)<\frac{2\gamma\sqrt{k\left(t\right)^3}}{\sqrt{k\left(t\right)}+2\zeta\,\gamma\sqrt{m}}\end{equation}\normalsize 
In order to switch the robot role between follower, i.e. compliant to the external force ($k_{min}$), and leader ($k_{max}$) \cite{Jarrasse2014}\cite{Gomes2018}, we propose a continuously differentiable scalar role factor $\alpha\left(t\right)\in\left[0, 1\right]$ and the following varying stiffness profile:\begin{equation}k\left(t\right)=k_{min}+\alpha\left(t\right)\left(k_{max}-k_{min}\right)\end{equation}
\setcounter{figure}{3}
\begin{figure*}[t]
\centering
\includegraphics[width=0.98\linewidth]{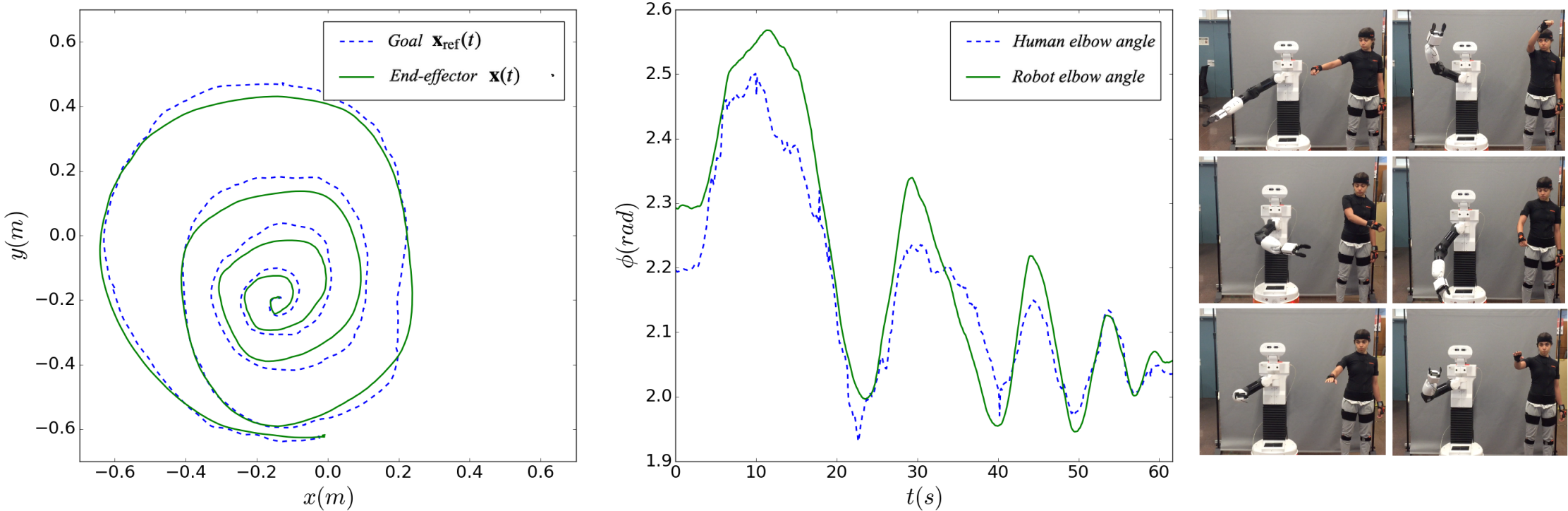}
\caption{From left to right: The operator hand reference (in dashed line) and the robot's end-effector (in continuous line) trajectories on the x-y plane; evolution over time of the operator and the robot elbow-wrist and elbow-shoulder segments angle $\phi$; finally, a series of snapshots of the experiment using the TIAGo robot and the motion capture system. The mean absolute error for the end-effector position is $11\;cm$ and $0.05\;rad$ for the elbow angle.}
\label{Fig5}
\vspace{-1.6em}
\end{figure*}
Role adaptation can be derived from the interaction force feedback. Experience of varying stiffness control suggests that continuous and smooth variations show no destabilization tendencies. We propose the following role factor profile:\begin{equation}\small\alpha\left(t\right)=\frac{1}{1+e^{-\left(a\psi\left(t\right)+b\right)}}\end{equation}\normalsize where $a$ and $b$ are design parameters and $\psi\left(t\right)\in\left[0,1\right]$ is a proposed interaction factor that varies according to the interaction force feedback. Note that higher values of $a$ give a faster transition between roles while $b$ determines the value of $\psi\left(t\right)$ at which the transition starts. We propose the following interaction factor dynamics:\begin{equation}\small\dt{\psi}\left(t\right)=\begin{cases} 
c^+\;\;\;\;\;\;\text{if}\quad \left\|\mathbf{F_{ext}}\left(t\right)\right\|> F_{thres}\;\;\text{and}\;\;\psi\neq 1\\
c^-\;\;\;\;\;\;\text{if}\quad \left\|\mathbf{F_{ext}}\left(t\right)\right\|\leq F_{thres}\;\;\text{and}\;\;\psi\neq 0\\
0\qquad\text{else}
\end{cases}\end{equation}\normalsize where $F_{thres}$ is the force threshold to consider physical human-robot interaction, and $c^-<0$ and $c^+>0$ are design parameters. Note that the values of $c^+$ and $c^-$ modulate the transition speed when switching from leader to follower and from follower to leader roles respectively. As a design guideline, for safety reasons it is important to achieve a fast stiff to compliant transition, but that is not the case for the opposite. Thus, high $c^+$ values are desirable but $c^-$ values should be kept relatively smaller in absolute value. 

From the first stability condition, since the damping profile is bounded, taking the least conservative constraints, we obtain $\gamma=2\zeta\sqrt{k_{min}}$. Given that all the varying parameters are bounded, we can determine an upper bound of the stiffness profile derivative, and a lower bound for the second stability condition. Thus, a sufficient stability condition is:\begin{equation}\frac{-a\cdot e^{-b}\left(k_{max}-k_{min}\right)c^-}{\left(1+e^{-b}\right)^2}<\frac{4\zeta\sqrt{k_{min}^3}}{1+4\zeta\sqrt{m}}\end{equation}

Tuning the parameters empirically, we have assigned $\zeta=1.1$, $m=1\;kg$, $k_{min}=10\;Nm^{-1}$, $k_{max}=500\;Nm^{-1}$ ($Nm\,rad^{-1}$ for the rotational components), $a=20$, $b=-5.5$, $c^-=-0.2$ and $c^+=1.5$ for all the DOF. By direct substitution, the sufficient stability condition holds. For filtering the noise in the interaction force feedback signal coming from the 6-axis force sensor we implemented a moving average filter \cite{Smith1997} of 25 samples (40 Hz sampling rate). An overview of the variable admittance controller can be seen in Figure \ref{Fig4}.

\setcounter{figure}{2}
\begin{figure}[h]
    \centering
    {\includegraphics[width=0.66\linewidth]{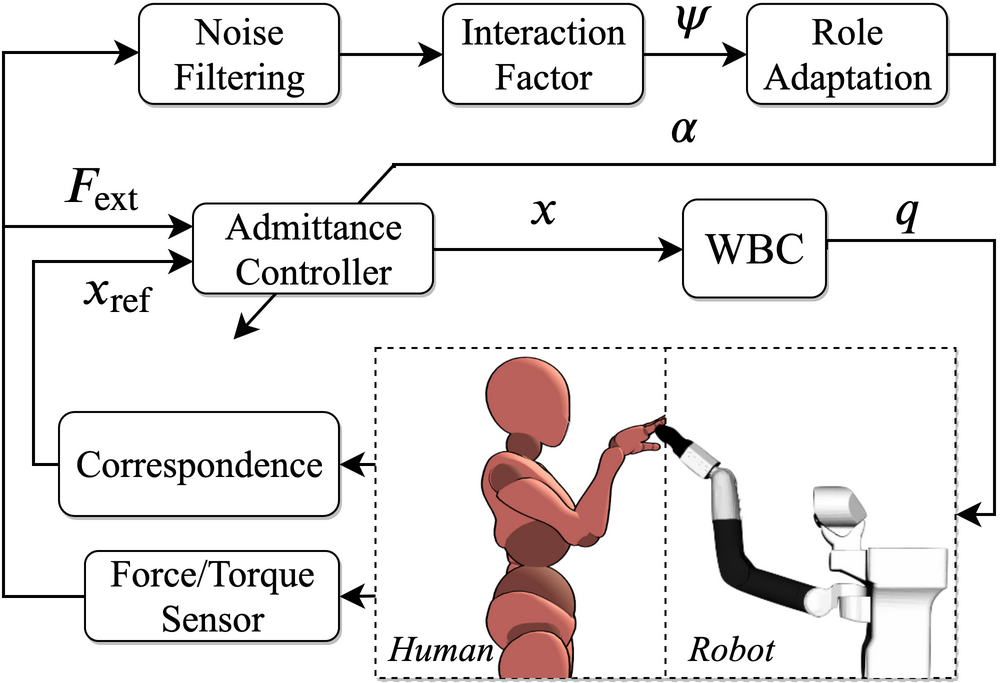}}
    \caption{Role adaptive admittance controller with human in the loop.}
    \label{Fig4}
    \vspace{-0.75em}
\end{figure}

\section{Experimental Results}
\label{experiments}

We carried out three different real-world experiments to validate the proposed approach, using the TIAGo robot, with 10 DOF excluding the head, and the Xsens motion capture system. The objective of the first experiment was to show the upper-body motion similarity when using the proposed solution for the correspondence problem and the WBC with the presented task hierarchy. In the second experiment we tested the mobile base imitation using the proposed algorithm for differential drive control. For the third experiment we evaluated the performance of the role adaptation mechanism and that the stability condition derived analytically is sufficient to ensure a stable behavior. Additionally, several demonstrations are included as a supplementary video. 

\subsection{Upper-body Motion Transfer}

The robot performed real-time imitation while the human operator described a spiral trajectory with the hand. To evaluate the motion similarity, we compared the trajectory described by the robot's end-effector and the evolution of the angle formed by the robot elbow-wrist and elbow-shoulder segments with the reference trajectories. Results are shown in Figure \ref{Fig5}. The obtained mean absolute error for the end-effector position is of $11\;cm$ and of $0.05\;rad$ for the elbow opening angle. As it can be seen, the robot is able to describe a spiral with the end-effector accurately while imitating the arm posture with its 7 DOF, proving a successful redundancy resolution. It can also be seen, from inspecting the results, that although a real-time imitation is achieved, the commanded motion is of an average speed of $11\;cm/s$. During the experiments, we observed that due to the robot's joint speed limits and own inertia, the operator movements should be limited to low speed motions.

\setcounter{figure}{5}
\begin{figure*}[t]
\centering
\includegraphics[width=0.99\linewidth]{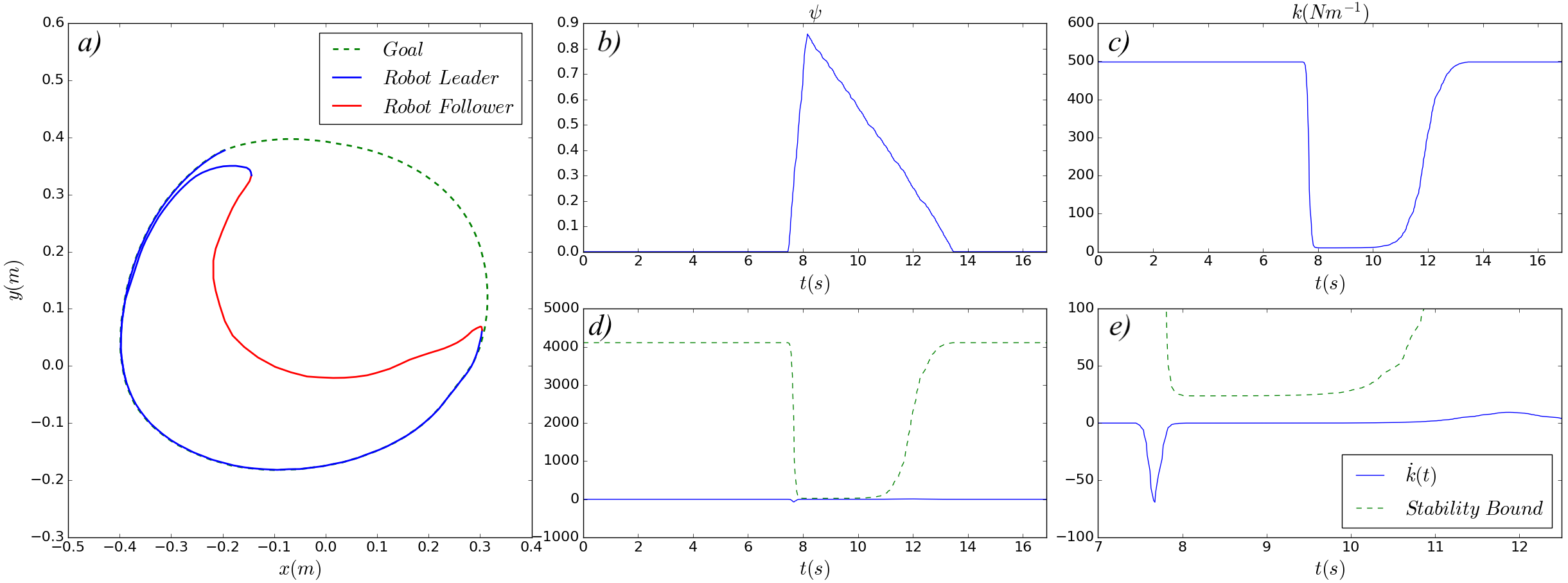}
\caption{(\textbf{a}) End-effector goal trajectory (dashed line). In blue, the trajectory described by the end-effector when the robot is playing the leader role (high stiffness). In red, the trajectory followed when the end-effector is grasped and the robot is playing the follower role (low stiffness). (\textbf{b}) Evolution of the interaction factor. (\textbf{c}) Stiffness profile. (\textbf{d}) Stability bound for the derivative of the stiffness profile (dashed line) and stiffness derivative evolution (continuous line). (\textbf{e}) A closer look at the area of the previous plot where the derivative and the stability bound reach the minimum difference.}
\label{Fig7}
\vspace{-1.5em}
\end{figure*}

\setcounter{figure}{4}
\begin{figure}[h]
    \centering
    {\includegraphics[width=0.97\linewidth]{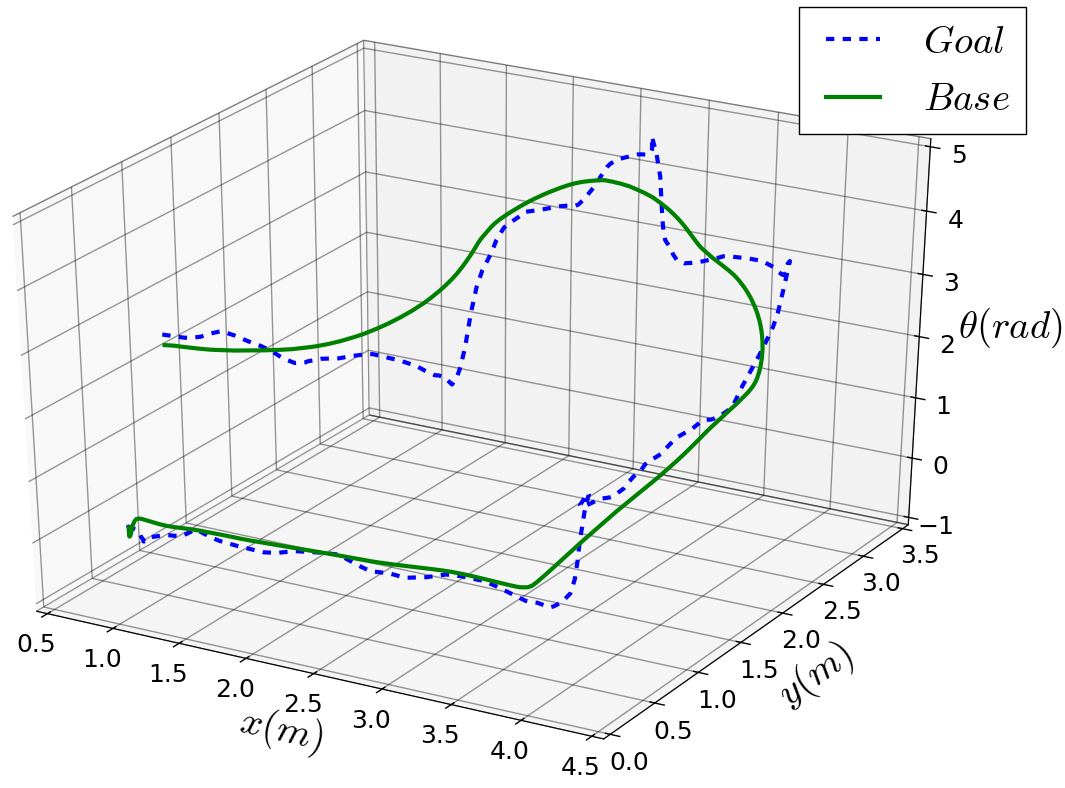}}
    \caption{Position ($x$ and $y$) and orientation ($\theta$) along a path described by the robot base (continuous line) and the goal trajectory (dashed line). The mean absolute error in position is $19\;cm$ and $0.31\;rad$ in orientation.}
    \label{Fig6}
    \vspace{-1.8em}
\end{figure}

\subsection{Base Motion Transfer}

The operator described a path with a series of turns while the robot moved in parallel. The obtained results are shown in Figure \ref{Fig6}. The mean absolute error in position is of $19\;cm$ and of $0.31\;rad$ in orientation. It can be observed that the robot motion is very similar to the reference trajectory. We are able to imitate the operator pose during the walking motion through velocity commands with the proposed algorithm. It should be remarked that the non-holonomic constraint does not apply to human walking motion. Therefore, in order to achieve a successful imitation the operator trajectory should not include side steps. However, when the non-holonomic constraint is not satisfied in the operator movement, for a sufficient large time, the base position and orientation always converge to the reference if it remains static.

\subsection{Role Adaptation}

A reference trajectory was commanded to the robot, while executing the motion, the end-effector is grasped by a person, displacing it from its goal trajectory. Then, it is released. The experiment results are shown in Figure \ref{Fig7}. The results show how, when the grasping occurs, the interaction factor starts to increase, while the stiffness rapidly decreases to switch the robot behavior from stiff to compliant. This allows to easily move the end-effector away from its commanded trajectory. When it is released, the interaction factor starts to decrease while the stiffness starts to restore its initial value and the robot end-effector position converges to the original trajectory. Note that when the stiffness is at its minimum value the difference between the stability bound and the stiffness profile derivative reaches its minimum value. Nevertheless, stability is fulfilled during the whole realization. No oscillations or unstable behavior were observed. 

\section{Conclusion}
\label{conclusion}

In this paper we have presented a human to robot whole-body motion transfer framework. Imitation, on the one hand, offers many advantages, not only because it is intuitive, but also because it allows to transfer human-like motion to the robot. On the other hand, it involves solving the correspondence problem. We present a novel general solution, that first, defines the equivalence between an arbitrary human body posture and the corresponding robot posture as a goal pose for a series of links in Cartesian space. Then, we propose a whole-body control scheme to find the robot configuration that attains simultaneous goals. By defining an adequate task hierarchy, we achieve an effective upper-body redundancy resolution. Furthermore, we have presented an algorithm that allows the robot differential drive base to imitate the human translation (through walking) motion. Finally, when a robot is operating in human-shared environments it is important to ensure safe human-robot interaction. However, achieving a compliant behavior and accurate position control are opposite objectives. We propose a novel variable admittance controller that allows continuous adaptation of the end-effector dynamics when physically interacting with a human by means of scalar role and interaction factors. For the proposed controller, we have derived analytically a state-independent and sufficient condition for ensuring stability. 

Experimental results show that an effective whole-body imitation is achieved in real-time. Moreover, the robot successfully adapts its role when physical interaction with a person occurs. Experimentation has shown that the main limiting factors preventing faster imitation are the robot's joint speed limits and inertia. Imitating translation, for differential-drive bases is limited because of the non-holonomic constraint.

This work has contributed a building block to a robotic system able to learn skills through demonstrations. The proposed approach to transfer human body motion to a mobile manipulator provides an easy way for a non-expert to teach a rough manipulation skill to a service or assistive robot. Afterwards, the robot would autonomously practice and improve the skill (e.g., its accuracy) through reinforcement learning \cite{Colome2020}. Future research towards learning dexterous manipulation skills will address the challenge of generalizing and adapting the learned motion when dealing with uncertainty. The combination of imitation learning and variable admittance control is a promising first step towards robots performing complex manipulation tasks in human-shared environments.

\bibliographystyle{IEEEtran}
\bibliography{Referencias}

\end{document}